\documentclass[sn-apa]{sn-jnl}
\usepackage{hhline}
\usepackage{fixltx2e}

\jyear{2023}%

\theoremstyle{thmstyleone}%
\theoremstyle{thmstyletwo}%
\usepackage[dvipsnames]{xcolor}
\theoremstyle{thmstylethree}%

\begin{document}

\title[Trie-NLG]{\textsc{Trie-NLG}: Trie Context Augmentation to Improve Personalized Query Auto-Completion for Short and Unseen Prefixes}

\author*[1]{\fnm{Kaushal Kumar} \sur{Maurya}}\email{cs18resch11003@iith.ac.in}

\author[1]{\fnm{Maunendra Sankar} \sur{Desarkar}}\email{maunendra@cse.iith.ac.in}

\author[2]{\fnm{Manish} \sur{Gupta}}\email{gmanish@microsoft.com}
\author[2]{\fnm{Puneet} \sur{Agrawal}}\email{punagr@microsoft.com}

\affil[1]{\orgname{IIT Hyderabad}, \orgaddress{\city{Hyderabad}, \postcode{502284}, \state{Telangana}, \country{India}}}

\affil[2]{\orgname{Microsoft India}, \orgaddress{\city{Hyderabad}, \postcode{500032}, \state{Telangana}, \country{India}}}

%\affil[3]{\orgdiv{Department}, \orgname{Organization}, \orgaddress{\street{Street}, \city{City}, \postcode{610101}, \state{State}, \country{Country}}}


\abstract{Query auto-completion (QAC) aims at suggesting plausible completions for a given query prefix. Traditionally, QAC systems have leveraged tries curated from historical query logs to suggest most popular completions. In this context, there are two specific scenarios that are difficult to handle for any QAC system:  short prefixes (which are inherently ambiguous) and unseen prefixes. Recently, personalized Natural Language Generation (NLG) models have been proposed to leverage previous session queries as context for addressing these two challenges. However, such NLG models suffer from two drawbacks: (1) some of the previous session queries could be noisy and irrelevant to the user intent for the current prefix, and (2) NLG models cannot directly incorporate historical query popularity. This motivates us to propose a novel NLG model for QAC, \textsc{Trie-NLG}, which jointly leverages popularity signals from trie and personalization signals from previous session queries. We train the \textsc{Trie-NLG} model by augmenting the prefix with rich context comprising of recent session queries and top trie completions. This simple modeling approach overcomes the limitations of trie-based and NLG-based approaches, and leads to state-of-the-art performance. We evaluate the \textsc{Trie-NLG} model using two large QAC datasets. On average, our model achieves huge $\sim$57\% and $\sim$14\% boost in MRR over the popular trie-based lookup and the strong BART-based baseline methods, respectively. We make our code publicly available\footnote{\url{https://github.com/kaushal0494/Trie-NLG}}}.


\maketitle

\section{Introduction}
Query formulation could be time-consuming for na\"ive users or users with complex information needs. Modern search engines, therefore, have a Query Auto-Completion (QAC) module to assist users in efficiently expressing their information need as a search query. The goal is to help users finish their search task faster by accurately understanding their query intent using the partially-typed prefix. While users type a partial search query (i.e., query prefix), the QAC system recommends a list of relevant complete queries (i.e., query auto-completions or suggestions).

Most of the popular search engines adopt a two-stage approach for QAC: \textit{candidate retrieval} and \textit{candidate ranking}~\citep{cai2016query}. A set of prefix-preserving suggestions is retrieved from a pool of complete candidate queries in the candidate retrieval stage\footnote{A small percent of suggestions are not prefix preserving; in this work, we focus on prefix-preserving suggestions only.}. Typically this is supported using a trie that records complete suggestions along with their historical popularity scores computed over a time window. Candidate retrieval could leverage various heuristics like historical candidate popularity, language or region-based affinity, freshness, etc. In the candidate ranking stage, these retrieved queries are ranked based on a larger list of features including popularity, the user's previous search intent, the user's profile, etc. Finally, top-N-ranked candidates are shown to the user. 

Although QAC has been studied for many decades, there are two major challenges yet to be solved. 
\begin{enumerate}
    \item \textbf{Short Prefixes:} High-quality completions for very short prefixes are the most desirable feature for any QAC system. But short prefixes are likely to have a huge candidate pool from the trie, and most of the QAC models return the most popular completions that may not be relevant.
    \item \textbf{Unseen prefixes:} Trie-based systems fail to provide recommendations for prefixes that have never been recorded previously, i.e., not a part of the query log. We refer to such prefixes as \textit{unseen} prefixes. 
\end{enumerate}

\begin{figure} [!htb]
% \captionsetup{font=scriptsize}
\centering
\includegraphics[width=\columnwidth]{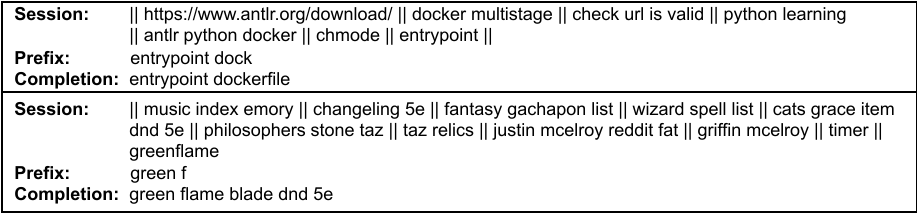}
\caption{Examples of session queries, prefix and completion. Queries are separated by `$\|$'.}
\label{fig:session_sample}
\end{figure}

To overcome these problems, more recently, seq2seq models have gained attention \citep{dehghani2017learning,mustar2020using,yin2020learning}. Besides the current prefix, these neural network-based natural language generation (NLG) models are more powerful because they can also utilize relevant session information to recommend personalized query completions. But even NLG models have the following drawbacks: (1) Unlike trie-based methods, NLG models cannot directly incorporate historical popularity which is a very important signal.  %(2) Traditionally, a session has been defined as a short time window (new session starts if consecutive queries have, say, at least 30 minutes of idle time between them). However, with increased levels of multi-tasking, sessions have become heterogeneous, diverse and dynamic. This makes it difficult to focus on session queries relevant to the current prefix~\citep{yin2020learning}. 
(2) With increased levels of multi-tasking, sessions have become heterogeneous, diverse and dynamic. This makes it difficult to focus on session queries relevant to the current prefix~\citep{yin2020learning}. A few such session examples are presented in Fig.~\ref{fig:session_sample}.
(3) Attention-based NLG methods which attempt to discover relevant session queries by computing similarity with prefix representations suffer when prefixes are too short. Misleading attention leads to poor completions. (4) As the unseen prefixes are typed rarely, corresponding session information may not be very relevant. 

NLG models can nicely capture the semantic relationships between existing session queries, prefix and completion. 
%In some ways they encode such semantic patterns. 
On the other hand, information in tries is like frequency-based high-confidence rules that capture relationships between prefix and completions in a syntactic manner. We hypothesize that jointly leveraging popularity signals from trie, and semantic and personalization signals from previous session queries using an NLG mechanism are essential for effective QAC. Based on this hypothesis, we propose a novel model for QAC, \textsc{Trie-NLG},  which uses a sequence-to-sequence Transformer architecture. To the best of our knowledge, such joint modeling of NLG techniques with popularity signals from trie for query auto-completion has not been studied in the literature.

%based pre-trained language model like BART~\citep{lewis2019bart} or T5~\citep{JMLR:v21:20-074}. 

Given a prefix, \textsc{Trie-NLG} first extracts up to top-$m$ most popular completions from the trie. We utilize a trie with around one billion suggestions constructed using 1.5 years of past query logs (Jul 2020 to Dec 2021) from Bing. For unseen prefixes, trie lookups lead to no (prefix-preserving) matches. To solve this problem, inspired by~\cite{mitra2015query}, we first index all suffix word n-grams from query logs into a suffix trie along with suffix popularity. We then lookup unseen prefixes against the suffix trie to extract top-$m$ most popular synthetic completions. These $m$ popularity-based completions, either from the main trie or from suffix trie, are augmented as extra context along with session queries and prefixes and passed as input to the seq2seq model. We hope that having additional knowledge from trie-lookup will enable the NLG model to retain/copy good quality completions along with the generation of the novel but relevant suggestions. 

Overall, our main contributions are as follows:
\begin{itemize}
    \item We motivate the need for incorporating both popularity signals from tries and personalization signals from previous session queries for effective QAC, especially for short and unseen prefixes.
    \item We propose a novel architecture, \textsc{Trie-NLG}, which consists of a seq2seq Transformer model trained using rich context comprising of recent session queries and top trie completions. To the best of our knowledge, this is the first attempt of trie knowledge augmentation in NLG models for personalized QAC.
    \item Our proposed model provides state-of-the-art performance on two real prefix-to-query click behaviour QAC datasets from Bing and AOL. We also perform several analyses including ablation studies to prove the robustness of the proposed model.
\end{itemize}

The rest of the paper is organized as follows. We discuss related work on traditional, learning-based, and language generation-based approaches for query auto-completion in Section~\ref{sec:relatedWork}. Next, we formally define the problem in Section~\ref{sec:problem}. 
%In Section~\ref{sec:datasets}, we present the dataset details, including data construction steps, and some critical observations. 
In Section~\ref{sec:methodology}, we present details of our proposed \textsc{Trie-NLG} modeling approach. We present dataset details, evaluation metrics, baselines, and implementation details in Section~\ref{sec:expts}. We present and analyze the results of different baselines and the proposed model in Section~\ref{sec:results}. We conclude with a brief summary in Section~\ref{sec:conclusions}.

\section{Related Work}
\label{sec:relatedWork}
In this section, we focus on three threads of related work for Query Auto-Completion (QAC), viz., traditional, learning-based, and language generation-based approaches.  

\subsection{Traditional Approaches for QAC}
Most of the traditional QAC systems leverage \textit{tries}~\citep{hsu2013space} which store historical co-occurrence statistics of prefix and complete query pairs. The most popular QAC approach using trie lookups is \textit{``Most Popular Completion"} (MPC; \cite{bar2011context}) which suggests top-N most popular (frequent) queries that start with the given prefix.~\citet{mitra2015query} extended this approach to generate candidates for rare prefixes using frequently observed query suffixes mined from historical search logs. On similar lines, other methods rely on term co-occurrence~\citep{huang2003relevant}, user click information~\citep{mei2008query}, clustering queries~\citep{sadikov2010clustering}, and using word level representations~\citep{bonchi2012efficient}. Some previous studies~\citep{bhatia2011query,maxwell2017large} also focused on modeling approaches when search logs are not available. 

\subsection{Learning-Based Approaches for QAC}
Query log-based approaches are usually context-agnostic and suffer from data sparsity issues. It is critical to leverage context for capturing personalized intent and behavior. To cope with these limitations, different sources of knowledge have been exploited in the candidate ranking stage with the learning-to-rank framework~\citep{wu2010adapting}. These additional signals include session information~\citep{bar2011context,jiang2014learning}, user behavior~\citep{hofmann2014eye,mitra2014user}, personalization~\citep{cai2014time,shokouhi2013learning} and time/popularity-sensitivity~\citep{shokouhi2012time}. Learning methods include LambdaMART~\citep{burges2010ranknet}, logistic regression~\citep{shokouhi2013learning}, convolutional neural network (CNN; \cite{mitra2015query}), deep learning based ranking model (DRM; \cite{zhou2018deep}), and eXtreme Multi-Label Ranking~\citep{yadav2021session}.
%explored end-to-end retrieve-and-rank framework of eXtreme Multi-Label Ranking (XMR) model for personalized QAC. 
These ranking models, however, fail to generate completions for unseen prefixes. Unlike these, we develop NLG models which capture personalization, learn contextual input representations, and provide completions even for unseen prefixes. 

\subsection{NLG-Based Approaches for QAC}
Recently, sequence-to-sequence language model-based approaches have also been tried for QAC~\citep{park2017neural,wang2018realtime}. Given a prefix and optionally personalization information, these models generate prefix-preserving completions. %For multiple generations, a beam search decoder is used. 
These models can generate completions for unseen prefixes.~\citet{wang2018realtime} use LSTM (Long Short-Term Memory networks) and GRU (Gated Recurrent Units) based character-level language model to generate completions.~\citet{dehghani2017learning} proposed GRUs with attention and copy mechanism to incorporate the most prominent part of the previous queries.~\citet{mustar2020using} and~\citet{yin2020learning} proposed  Transformer~\citep{vaswani2017attention} based  models.~\citet{yin2020learning}'s approach requires additional browsed item information and also needs CTR values as labels to train the model. Moreover, these generation models still fail to generate good completions for short and rare prefixes. Unlike these methods, we encode additional trie context along with session in the NLG model which leads to more meaningful completions for short, rare, and unseen query prefixes. Note that in our case, additional context is obtained from tries that are a part of any QAC system.

\section{Problem Formulation}
\label{sec:problem}
Consider a user $u$ whose previous $n$ queries (earliest to latest order) in the current session $s$ are $q_1, q_2,$ \dots, $q_n$. The user is typing the current query $q$, where $p$ is the query prefix typed so far.  Additionally, there are \textit{up to} $m$ candidate query completions (top-ranked to low-ranked order) $c_1, c_2,$ \dots, $c_m$ available as additional context $e$ from a trie. We aim to generate top-$N$ query completions conditioned on current query prefix $p$, additional trie context $e$, and session information $s$. Mathematically, the task can be formulated as learning a model with parameters $\theta$ such that the probability of generating query $q$, $P_{\theta}(q \lvert p; c_1, c_2, \dots, c_m;q_1, q_2, \dots, q_n)$, is maximized. Here we consider the value of $N$ is equal to 8, i.e., the number of auto-completions is 8.

\section{Methodology}
\label{sec:methodology}

\begin{figure}[!htb]
\centering
\includegraphics[width=\columnwidth]{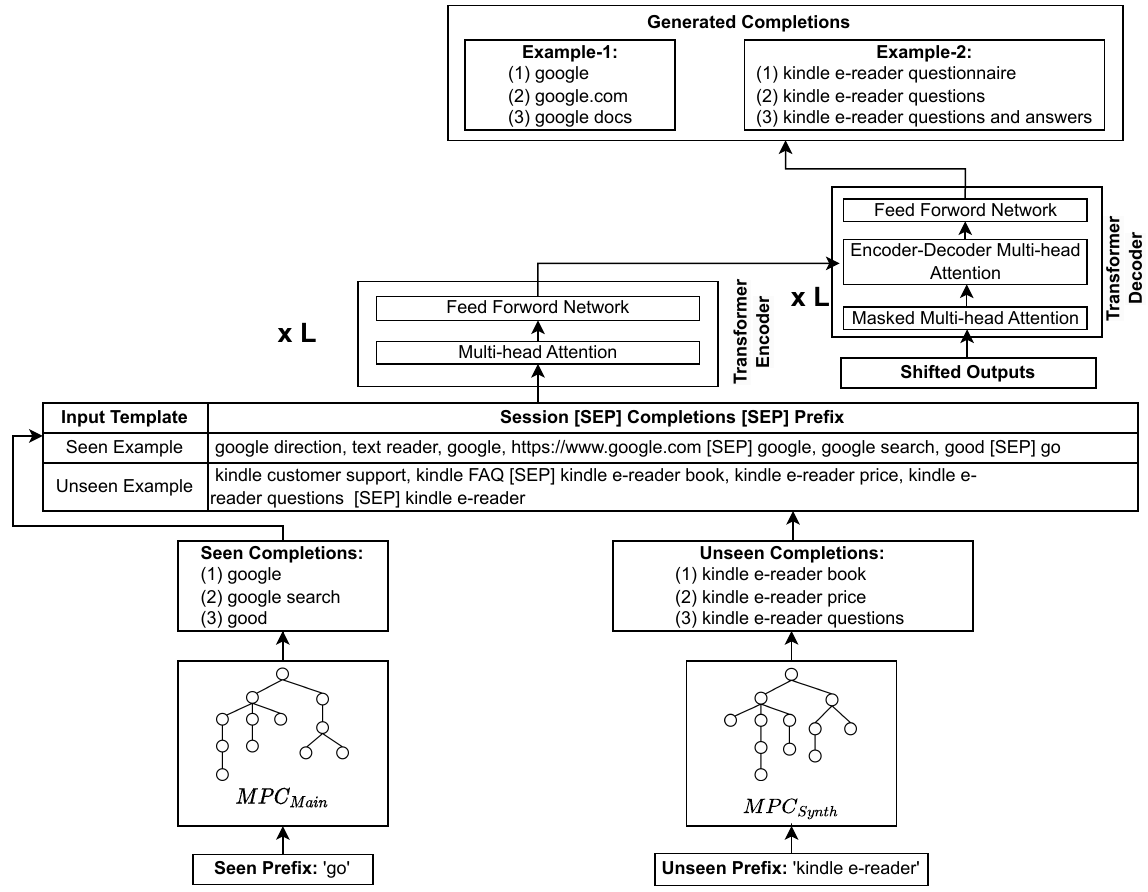}
\caption{An overview of the proposed \textsc{Trie-NLG} model}
% \caption{ Detailed illustration of proposed \textsc{Trie-NLG} model with running examples}} 
\label{fig:detailed_trie_nlg}
\end{figure}

The proposed \textsc{Trie-NLG} model extracts a few completions from the trie and augments them as part of the input to an NLG model. For a given prefix, up to top-$m$ completions are extracted as additional context from the trie using MPC. Those prefixes for which completions can be obtained from the MPC are called \textit{Seen} prefixes, while those for which completions are not present are called \textit{Unseen} prefixes. 
%More discussion on seen and unseen prefixes is presented in Section \ref{sec:datasets}. 
For seen prefixes, we leverage the main trie, and for unseen prefixes, we leverage a new trie called as suffix trie. These suggestions from the main or suffix trie are augmented with previous queries in the session and the current prefix, and passed as input to the NLG model to generate accurate completions. Fig.~\ref{fig:detailed_trie_nlg}
%~\ref{fig:trie_nlg_qac} and 
illustrates the architecture of the proposed model. To enable a concrete understanding of the proposed model, we consider two running examples. For simplicity, we consider only the prefix and ground truth completion in the \textit{$\langle$prefix, completion$\rangle$} template. Example-1: \textit{$\langle$go, google.com$\rangle$} and Example-2: \textit{$\langle$kindle e-reader, kindle e-reader questionnaire$\rangle$}.

% The proposed \textsc{Trie-NLG} model extracts a few completions from the trie and augments them as part of the input to an NLG model. For a given prefix, up to top-$m$ completions are extracted as additional context from the trie using MPC. Those prefixes for which completions can be obtained from trie called as \textit{Seen} prefix for those completions are not present called as \textit{Unseen} prefix. More discussion on seen and unseen prefixes are presented in section \ref{sec:datasets}. For seen prefixes, we leverage the main trie. For unseen prefixes, we leverage the suffix trie.  These suggestions from main/suffix trie are augmented with previous queries in the session and current prefix and passed as input to the NLG model to generate accurate completions. Fig.~\ref{fig:trie_nlg_qac} and \ref{fig:detaied_trie_nlg} illustrate the architecture of the proposed model. \textcolor{brown}{ To enable a concrete understanding of the proposed model we have considered two running examples, for simplicity, we consider only prefix and ground truth completion in \textit{$<$prefix, completion$>$} template. Example-1: \textit{$<$go, google.com$>$} and example-2: \textit{$<$kindle e-reader, kindle e-reader questionnaire$>$}.}

% \begin{figure} [!htb]
% % \captionsetup{font=scriptsize}
% \centering
% \includegraphics[width=0.6\columnwidth]{figures/Trie_QAC3.pdf}
% \caption{An overview of the proposed \textsc{Trie-NLG} model} 
% \label{fig:trie_nlg_qac}
% \end{figure}

\subsection{Trie Context Extraction (MPC\texorpdfstring{\textsubscript{Main}}{Main})}

To extend the context associated with seen prefixes, top-ranked completions are extracted from the main trie which has been created using 1.5 years' worth of Bing query logs. Given a prefix $p$,  MPC\textsubscript{Main} provides up to $m$ completions $\langle c_1, c_2,$ \dots, $c_m\rangle$. In case the prefix is not present in the trie, the lookup will return no responses. For our running example-1, for the prefix \textit{go}, MPC\textsubscript{Main} returns three completions: \textit{google}, \textit{google.com}, and \textit{good}. However, for running example-2, no completions are obtained from MPC\textsubscript{Main} for the prefix \textit{kindle e-reader}. The prefix \textit{go} is referred to as \textit{seen} prefix, while \textit{kindle e-reader} is referred as \textit{unseen} prefix.
% \begin{equation}
%     \langle c_1, c_2, \dots, c_m \rangle  = \textrm{MPC}\textsubscript{Main}(p)
% \end{equation}

\subsection{Synthetic Context Extraction (MPC\texorpdfstring{\textsubscript{Synth}}{Synth})}
\label{synthmpc}

For unseen prefixes, the main trie fails to provide any completions. In such cases, we make use of another trie called the suffix trie which is created by indexing all suffix word n-grams from query logs along with suffix popularity. Since the suffix word n-grams may not be actual queries, we call them synthetic suggestions. Formally, if a query contains $n$ words $w_1,...w_n$, its substrings $\langle w_i,...w_n\rangle$ for each $i\in\{1,\cdots,n\}$ is a suffix of it. These suffixes are organized in another trie called the suffix trie. For a given unseen prefix $p_u$, if the main trie fails to return any suggested completions, we look up the suffix trie for suggestions. Matching unseen prefixes in suffix trie enables locating queries that contain the current prefix $p_u$ as an intermediate word. The words surrounding the prefix provide additional context that is useful to generate the final query.

For example, given a query ``university of west florida'', the suffix trie will store synthetic suggestions like ``florida'', ``west florida'' and ``of west florida''. We lookup unseen prefixes against the suffix trie to extract top-$m$ most popular synthetic yet useful completions. Note that these lookups still attempt to match the unseen prefix with prefixes of suffixes indexed in the suffix trie. In this way, we will be able to obtain completions for unseen prefixes which can not be obtained from MPC\textsubscript{Main}. We refer to this method as MPC\textsubscript{Synth}. Although this idea is similar to one described by~\cite{mitra2015query}, unlike them, we consider the whole prefix and not only \textit{end-term} of the prefix. If a prefix has multiple words, the last partial word is the end-term. The whole prefix has more meaningful contextual representation than end-term representation which leads to more accurate completions. 
%Illustration of MPC\textsubscript{Synth} model is presented in Fig.~\ref{fig:sample_mpcsynth}. 
For running example-2, MPC\textsubscript{Synth} returns three completions for the unseen prefix \textit{kindle e-reader}: \textit{kindle e-reader book}, \textit{kindle e-reader price}, and \textit{kindle e-reader questions.}
% Formally, given prefix $p$, MPC\textsubscript{Synth} provides up to $m$ completions $\langle c_1, c_2,$ \dots, $c_m\rangle$.

% \begin{equation}
%   \langle  c_1, c_2, \dots, c_m \rangle  = \textrm{MPC}\textsubscript{Synth}(p)
% \end{equation}

% \begin{figure} [!htb]
% % \captionsetup{font=scriptsize}
% \centering
% \includegraphics[width=\columnwidth]{figures/MCP_synth_sample.pdf}
% \textcolor{brown}{
% \caption{ Creation of MPC\textsubscript{Synth} model. Additionally, we have also shown toy completion examples of a prefix in different scenarios.}} 
% \label{fig:sample_mpcsynth}
% \end{figure}

\subsection{Context Augmentations in NLG}

After obtaining trie suggestions, each data point consists of the session information ($s$), additional trie context ($e$), prefix ($p$), and the corresponding completion ($q$). We consider an Encoder-Decoder-based NLG model that takes the triplet $\langle s, e, p\rangle$ as input and attempts to generate the complete query ($q$). The input is provided to the model as a text sequence, where each element of the triplet is separated by a special token [SEP]. Trie context, i.e., top-m candidate completions are obtained from MPC\textsubscript{Main} or MPC\textsubscript{Synth}. %These two types of prefix sets are mutually exclusive. 
During model training, the input triplet is first fed through the encoder to obtain a contextual representation. These contextual representations are semantic encodings of $\langle s, e, p\rangle$, which is key for the model's performance, particularly for short and unseen prefixes. Then, this contextual representation is passed through the decoder to generate top-N completions.

Relevant contextual suggestions from tries help the model with additional input that can guide the generation process. As typically the queries in a user session are often correlated in terms of the user's information need, the session context helps the model in understanding the user's current requirement. On the other hand, through the suggestions from the trie which is backed by historical query logs, a global perspective of the prefix and its possible completions preferred by a large user base can be obtained. The model thereby gets to see a local (concerning the user) as well as a global (concerning a large user pool) perspective surrounding the current prefix, and can appropriately utilize these inputs through language models pre-trained on large general-purpose corpora that understand semantic and syntactic aspects of natural language text. We hypothesize that the combination of these input and modeling choices makes the model superior for the target QAC task. 
%reduces the diversity and dynamic problem and provides a rich contextual representation of the input. 

The model is trained to maximize the probability of ground truth token sequence with maximum likelihood estimation (MLE). So, the following loss function is minimized: 
\begin{equation}
    L=- \sum_{i=1}^{D}\sum_{t=1}^{\lvert y^i\lvert}log P_t^i(\hat{y}_t^i \lvert \hat{y}_{0:t-1}^i;p;e;s)
\end{equation}
where $D$ is the training dataset size, $\lvert y^i\lvert$ is the length of the $i$-th ground-truth query, $\hat{y}_{t}$ denotes token generated at time step $t$. $P_t^i$ is the prediction probability distribution at $t$-th decoding step to generate the next token conditioned on previously generated tokens, prefix, trie context and session. The top-N completions are generated using beam search. For both running examples, both MPC\textsubscript{Main} and MPC\textsubscript{Synth} failed to produce the correct ground truth completion. However, with context augmentation in NLG, it can be observed that for the prefix \textit{go}, the final completion includes the correct completion \textit{google.com}. Similarly, the prefix \textit{kindle e-reader} has the correct completion \textit{kindle e-reader questionnaire.} This demonstrates the effectiveness of augmenting additional context in the NLG for QAC systems.    

\section{Datasets and Experimental Setup}
\label{sec:expts}
In this section, first, we provide a detailed overview of the dataset along with analyses then we compare the performance of the proposed model with multiple baseline models for the query auto-completion (QAC) task. Finally, We also provide a detailed analysis of the results from multiple perspectives  and report ablation studies and case studies. %First, we explain the different evaluation metrics and baselines used in the experimental evaluation.

\subsection{Datasets and Analysis}
\label{sec:datasets}

\setlength{\tabcolsep}{1.75pt}
\begin{table*}[!htb]
% \captionsetup{font=scriptsize}
\footnotesize
  \begin{center}
    \begin{tabular}{|l|c|c|c|c|c|c|c|c|c|} 
    \hline
   \textbf{Char} & \multicolumn{3}{c|}{\textbf{Train}} & \multicolumn{3}{c|}{\textbf{Validation}} &
    \multicolumn{3}{c|}{\textbf{Test}}\\ 
\hhline{|~|---------|}
      \textbf{Length} &\textbf{Total} & \textbf{Seen} & \textbf{Unseen} & \textbf{Total} & \textbf{Seen} & \textbf{Unseen} & \textbf{Total} & \textbf{Seen} & \textbf{Unseen}\\
      \hline
      \hline
       Total &20.40M&17.86M&2.54M&100K&92.43K&7.57K&100K&92.80K&7.20K\\ 
       $\textrm{[1-5]}$&9.10M&8.80M&0.30M&40.68K&40.39K&0.29K&40.46K&40.19K&0.27K\\ 
       $\textrm{[6-10]}$&4.30M&4.10M&0.20M&21.40K&21.07K&0.33K&21.62K&21.24K&0.38K\\ $\textrm{10+}$&7.00M&4.96M&2.04M&37.92K&30.97K&6.95K&37.92K&31.37K&6.55K\\ 
       \hline
    \end{tabular}
    \caption{Prefix distribution statistics for Bing dataset with prefix character length. `M' and `K' indicate that the value is in the order of millions and thousands respectively.}
    \label{tab:prefix_dist_bing}
  \end{center}
\end{table*}

\begin{table*}[!htb]
% \captionsetup{font=scriptsize}
\footnotesize
  \begin{center}
    \begin{tabular}{|l|c|c|c|c|c|c|c|c|c|} 
    \hline
   \textbf{Char} & \multicolumn{3}{c|}{\textbf{Train}} & \multicolumn{3}{c|}{\textbf{Validation}} &
    \multicolumn{3}{c|}{\textbf{Test}}\\ 
\hhline{|~|---------|}
      \textbf{Length} &\textbf{Total} & \textbf{Seen} & \textbf{Unseen} & \textbf{Total} & \textbf{Seen} & \textbf{Unseen} & \textbf{Total} & \textbf{Seen} & \textbf{Unseen}\\
      \hline
      \hline
       Total &3.91M&3.47M&0.44M&100K&88.73K&11.27K&100K&88.69K&11.31K\\ 
       $\textrm{[1-5]}$&1.42M&1.42M&0.00M&35.55K&35.54K&0.01K&36.59K&36.56K&0.03K\\ 
       $\textrm{[6-10]}$&1.15M&1.11M&0.04M&29.53K&28.67K&0.86K&29.51K&28.61K&0.90K\\ $\textrm{10+}$&1.34M&0.94M&0.40M&34.92K&24.52K&10.40K&33.90K&23.52K&10.38K\\ 
       \hline
    \end{tabular}
    \caption{Prefix distribution statistics for AOL dataset with prefix character length. `M' and `K' indicate that the value is in the order of millions and thousands respectively.}
    \label{tab:prefix_dist_aol}
  \end{center}
\end{table*}

\setlength{\tabcolsep}{4pt}

In this subsection, we present the dataset details, including data construction steps, and some critical observations. 
We use two datasets: (1) Bing query log and (2) AOL public query log~\citep{pass2006picture}. The Bing dataset covers 9.08M users, while the AOL dataset corresponds to 0.50M users.

%how the AOL dataset has been prepared, Introduction to the datasets and Dataset statistics
The raw AOL query log consists of a sequence of queries entered by the users along with time-stamp details. We first pre-process the dataset by lowercasing all the queries, removing duplicate and single character queries, and removing queries with dominating ($>$50\%) number of non-alphanumerics. Following previous studies~\citep{sordoni2015hierarchical,yadav2021session}, we split the sequence of queries into sessions with at least 30 minutes of idle time between two consecutive queries. We only retain those sessions which have at least two queries. Next, for a given session with queries (in earliest to latest order) $(q_1, q_2,$ \dots, $q_{n}$, $q_{n+1})$ we create a triplet $r=\langle(q_1, q_2, \dots,q_{n}), p_{n+1}, q_{n+1}\rangle$ where $p_{n+1}$ is a sampled prefix of query $q_{n+1}$. Sampling follows an exponential distribution favoring shorter prefixes. Each such triplet is a data point for modeling where input is all session queries except the last one, additional trie context, and prefix $p_{n+1}$. Ground-truth output is the last session query $q_{n+1}$. 

Unlike the AOL dataset, where the prefix-to-query information is not explicitly available, and the prefixes are synthetically created by splitting a full query, the Bing dataset consists of real prefixes. Each example of the dataset consists of the user's session information $s$, current real prefix $p_{n+1}$, and real clicked completed query $q_{n+1}$. The dataset is obtained by considering only those cases where there were at least one past query in the user session, and the user has set their primary language as English.      

Each dataset has two splits: \textit{Seen Dataset} and \textit{Unseen Dataset}. To obtain these splits, we use a trie with around one billion suggestions constructed using 1.5 years of past Bing query logs (Jul 2020 to Dec 2021). For a given prefix, if the trie contains at least one completion then the prefix is called a \textit{Seen Prefix}. Else, it is called an \textit{Unseen Prefix}. The set of all Seen Prefixes (along with other search log attributes) is referred to as \textit{Seen Dataset} and the set of all Unseen Prefixes is called \textit{Unseen Dataset}. Statistics of Bing and  AOL datasets shown in Tables~\ref{tab:prefix_dist_bing} and~\ref{tab:prefix_dist_aol} respectively indicate that there are 12.4\% and 11.4\% unseen prefixes in the training part of Bing and AOL datasets respectively. The traditional MPC baseline leads to zero suggestions for such prefixes.
The AOL and BING datasets each contain three sub-splits, namely train, validation, and test, which are based on temporal information and are applicable to both seen and unseen datasets as well. The timestamps for the train, validation, and test datasets are denoted as $t_{train}$, $t_{validation}$, and $t_{test}$, respectively. It is stipulated that $t_{train}$ is the least recent of the three timestamps, and $t_{validation}$ and $t_{test}$ are of increasing temporal proximity, with $t_{test}$ being the most recent of the three.

% \begin{table}[!htb]
% \captionsetup{font=scriptsize}
% \centering
% \scriptsize
% \begin{tabular}{|l|c|c|c|c|c|c|}
% \hline
% & \multicolumn{3}{c|}{\textbf{Bing}} & \multicolumn{3}{c|}{\textbf{AOL}}\\
% \cline{2-7} 
%  & \textbf{Train} & \textbf{Val} & \textbf{Test} & \textbf{Train}  & \textbf{Val} & \textbf{Test}\\
% \hline
% \hline
% Total \#Prefixes & 20.4M & 100K & 100K & 3.91M & 100K  & 100K \\
% Seen \#Prefixes& 17.86M & 92.43K  & 92.80K  & 3.47M  & 88.73K  & 88.69K \\
% Unseen \#Prefixes& 2.54M   & 7.57K & 7.20K   & 0.44M  & 11.27K & 11.31K\\
% \#Unique Users & 9.08M & 88.59K & 89.66K &  0.50M  & 76.47K & 76.33K\\
% %Log Time-Stamp & Mar'20-Feb'21 & Mar'20-May'20 & Mar'20-May'20 & Mar'06-May'06 & Mar'06-May'06 & Mar'06-Apr'06 \\
% \hline
% \end{tabular}
% \caption{Statistics of Bing and AOL datasets}
% \label{tab:data_stats}
% \end{table}

To analyze the accuracy of various methods for different slices of the dataset, we created three prefix-length buckets: between 1 to 5 characters, 6 to 10 characters, and greater than 10 characters. Prefix distribution statistics with respect to each bucket are also reported in Tables~\ref{tab:prefix_dist_bing} and~\ref{tab:prefix_dist_aol} for Bing and AOL datasets, respectively. We observe that $\sim$45\% and $\sim$36\% of the prefixes from the training datasets have lengths less than 6 for Bing and AOL, respectively. This indicates the dominance of short prefixes and necessitates the design of better modeling techniques. An approach that provides additional context (as we propose) is promising. We also observe that $\sim$80\% and $\sim$91\% of the unseen train dataset have prefix lengths 10+ characters for Bing and AOL respectively, which is another reason for them being less popular. The addition of more relevant context from tries may lead to better completions in such scenarios.  We also observe that the average number of queries in a session for Bing is 5.2 and for AOL is 2.4. This provides diverse personalization contexts to better judge the applicability and usefulness of the models. Overall, unseen and short prefixes in QAC are frequent and challenging problems.

\subsection{Evaluation Metrics}
We evaluate all the baselines and proposed model with three evaluation metrics.  To cover multiple aspects of the evaluation, we use both ranking-oriented metrics (MRR) and metrics to identify the quality of the generated sequence (BLEU and BLEU\textsubscript{RR}). %Additionally, we also use APPG metric directly from Microsoft's evaluation suite. 
% Details of the metrics are as follows.

% \noindent\underline{Generation-oriented Metric}
\begin{enumerate}
    \item \textbf{Bilingual Evaluation Understudy (BLEU):} It is a popular metric used for multiple NLG tasks. For our experiments, BLEU evaluates the degree of lexical match between the ground-truth complete query and the first ranked generated query.

% \noindent\underline{Ranking-oriented Metrics}

\item \textbf{Mean Reciprocal Rank (MRR):} MRR is one of the most popular metrics for evaluating ranking systems. MRR score is calculated as
\begin{equation}
    \textrm{MRR} =\frac{1}{D_{ts}}\sum_{i=1}^{D_{ts}}\frac{1}{r_i}
\end{equation}
Here, $D_{ts}$ is the size of the test dataset and $r_i$ is the rank of the ground-truth complete query in the generated rank list for the $i$\textsuperscript{th} input. If the ground-truth complete query is not in the generated rank list, then $r_i$ is set to $\infty$. %MRR scores return a value between $[0,1]$.

\item \textbf{BLEU Reciprocal Rank (BLEU\textsubscript{RR}; \cite{yadav2021session}}): It is defined as the reciprocal rank weighted average of BLEU score between the ground-truth query and generated completions.
\begin{equation}
    \textrm{BLEU}_{RR} =  \frac{1}{D_{ts}}\sum_{i=1}^{D_{ts}}\frac{\sum_{j=1}^{N} \frac{1}{j}\textrm{BLEU}(q, q'_{i,j}) }{\sum_{j=1}^{N}\frac{1}{j}}
\end{equation}
 where $q$ is the ground-truth complete query and $q'_{i,j}$ is the $j$-th generated completion for the $i$-th test example.
 \end{enumerate}
% \comr{\textbf{ROGUE?}}
%   \item \textbf{Average Prefix Level Prediction Gain (APPG):} This is a weighted MRR metric where weights are a function of query prefix length. This is one of the official metrics used to evaluate Microsoft's QAC production models. The details of the metric are not available publicly. We use the official script from Microsoft to compute the evaluation scores. 

\begin{sidewaystable}
\sidewaystablefn%
\footnotesize  
        \begin{minipage}{\textheight}
\centering
    \begin{tabular}{|l||c|c|c||c|c|c||c|c|c|} 
    \hline
     & \multicolumn{3}{c||}{ \textbf{Seen + Unseen Dataset}} & \multicolumn{3}{c||}{\textbf{Seen Dataset}} & \multicolumn{3}{c|}{\textbf{Unseen Dataset}} \\
    \hline
    \textbf{Models} &\textbf{$\Delta$MRR} & \textbf{$\Delta$BLEU\textsubscript{RR}} & \textbf{$\Delta$BLEU} & \textbf{$\Delta$MRR} & \textbf{$\Delta$BLEU\textsubscript{RR}}  & \textbf{$\Delta$BLEU} & \textbf{$\Delta$MRR} & \textbf{$\Delta$BLEU\textsubscript{RR}}  & \textbf{$\Delta$BLEU} \\
    \hline
    \hline
    MPC\textsubscript{Train} &-7.90&-19.87&-24.64&0.00&0.00&0.00& - & - & - \\
    MPC\textsubscript{Main}  &-0.11&36.61&26.38&8.49&70.38&63.68& - & -  & -\\
    MPC\textsubscript{Main} + MPC\textsubscript{Synth}  &7.78&56.42&47.01&8.49&70.38&63.68&0.00&0.00&0.00\\
    GRM &-1.43&-5.32&-5.51&4.30&19.02&16.58& - & -  & -\\
    Seq2Seq LSTM &9.61&39.30&56.78&10.24&44.02&72.20&5.34&16.61&91.45\\
    Seq2Seq Transformer &15.74&54.22&76.16&18.24&55.07&82.46&10.53&24.20&99.29\\
    T5 &20.78&61.81&78.70&20.76&70.83&85.13&21.60&27.33&103.98\\
    BART &36.73&73.47&91.09&36.95&84.51&100.47&34.53&29.03&110.35\\
    BART + ITC &31.66&71.43&88.72&31.58&81.97&96.84&33.05&29.12&111.00\\
    BART + MPC\textsubscript{Main}  &54.12&86.77&110.78&56.14&101.35&129.00&34.10&28.04&110.80\\
    \textsc{Trie-NLG}  & \textbf{56.78} & \textbf{88.26} & \textbf{114.52} & \textbf{56.56} & \textbf{101.99 }& \textbf{130.04} & \textbf{59.74} & \textbf{33.02}  & \textbf{123.07} \\
    \hline
    \end{tabular}
    \caption{Results of the models on Bing dataset. The reported scores are percentage (\%) improvements over MPC\textsubscript{Train} + MPC\textsubscript{Synth} baseline. `-' indicates no completions are retrieved/generated for the model. Here we consider up to 3 completions as additional context from MPC\textsubscript{Main} or MPC\textsubscript{Synth}. GRM is a ranking model based on clicked queries. As the Unseen dataset does not have click information, GRM models cannot be built.} 
    \label{tab:results_Bing}
    \end{minipage}
    \begin{minipage}{\textheight}
    \centering
    \begin{tabular}{|l||c|c|c||c|c|c||c|c|c|} 
    \hline
     & \multicolumn{3}{c||}{ \textbf{Seen + Unseen}} & \multicolumn{3}{c||}{\textbf{Seen Dataset}} & \multicolumn{3}{c|}{\textbf{Unseen Dataset}} \\
    \hline
    \textbf{Models} &\textbf{$\Delta$MRR} & \textbf{$\Delta$BLEU\textsubscript{RR}} & \textbf{$\Delta$BLEU} & \textbf{$\Delta$MRR} & \textbf{$\Delta$BLEU\textsubscript{RR}}  & \textbf{$\Delta$BLEU} & \textbf{$\Delta$MRR} & \textbf{$\Delta$BLEU\textsubscript{RR}}  & \textbf{$\Delta$BLEU} \\
    \hline
    \hline
    MPC\textsubscript{Train} & -34.18 & -55.37 & -61.97 & 0.00 & 0.00 & 0.00 & - & - & -
 \\
    %MPC\textsubscript{Train} + MPC\textsubscript{Synth} &31.3&12.0&40.10&23.2&6.1&19.49& \textbf{95.2} & \textbf{58.7}& \textbf{95.66}\\
    MPC\textsubscript{Main} & -37.06 & -18.18 & -39.05 & -4.31 & 83.60 & 52.64 & - & - & -
 \\
    MPC\textsubscript{Main}+MPC\textsubscript{Synth} &-2.87 & 37.19 & 19.10 & -4.31 & 83.60 & 52.64 & \textbf{0.00} & \textbf{0.00} & \textbf{0.00}\\
    GRM & -30.35 & -39.66 & -48.40 & 6.03 & 36.06 & 19.70 & - & - & - \\
    Seq2Seq LSTM &40.25 & 21.48 & 28.25 & 87.93 & 111.47 & 152.23 & -50.52 & -50.08 & -27.68\\
    Seq2Seq Transformer & 45.04 & 38.84 & 43.39 & 91.81 & 142.62 & 165.72 & -45.48 & -44.97 & -23.03\\
    T5 & 53.67 & 43.80 & 48.70 & 100.86 & 149.18 & 174.08 & -37.5 & -41.05 & -19.31\\
    BART & 65.81 & 51.23 & 54.33 & 116.81 & 163.93 & 185.47 & -32.03 & -39.01 & -16.84\\
    BART+ITC &61.98 & 51.23 & 53.49 & 111.63 & 163.93 & 183.68 & -33.29 & -38.84 & -17.16\\
    BART+MPC\textsubscript{Main}  & 69.96 & 53.71 & 55.81 & 123.27 & 168.85 & \textbf{190.30} & -32.66 & -39.35 & -17.19\\
    \textsc{Trie-NLG} & \textbf{80.51} & \textbf{59.50} & \textbf{66.15} & \textbf{124.56} & \textbf{170.49} & \textbf{190.20} & -3.25 & -29.81 & -1.08\\
    \hline
    \end{tabular}
    \caption{Results of the models on AOL dataset. The reported scores are percentage (\%) improvements over MPC\textsubscript{Train} + MPC\textsubscript{Synth} baseline. `-' indicates no completions are retrieved/generated for the model. Here we consider up to 3 completions as additional context from MPC\textsubscript{Main} or MPC\textsubscript{Synth}. GRM is a ranking model based on clicked queries. As the Unseen dataset does not have click information, GRM models cannot be built. 
    }
    \label{tab:results_AOL_relative}
    \end{minipage}
\end{sidewaystable}

\setlength{\tabcolsep}{1pt}
\begin{table}[!htb]
% \scriptsize 
\footnotesize
\centering
        \begin{tabular}{|l||c|c|c||c|c|c||c|c|c|} 
    \hline
     & \multicolumn{3}{c||}{ \textbf{Seen + Unseen Dataset}} & \multicolumn{3}{c||}{\textbf{Seen Dataset}} & \multicolumn{3}{c|}{\textbf{Unseen Dataset}} \\
    \hline
    \textbf{Models} &\textbf{MRR} & \textbf{BLEU\textsubscript{RR}} & \textbf{BLEU} & \textbf{MRR} & \textbf{BLEU\textsubscript{RR}}  & \textbf{BLEU} & \textbf{MRR} & \textbf{BLEU\textsubscript{RR}}  & \textbf{BLEU} \\
    \hline
    \hline
    MPC\textsubscript{Train} &20.6&5.4&15.25&23.2&6.1&19.49& - & - & - \\
    MPC\textsubscript{Train} + MPC\textsubscript{Synth} &31.3&12.0&40.10&23.2&6.1&19.49& \textbf{95.2} & \textbf{58.7}& \textbf{95.66}\\
    MPC\textsubscript{Main} &19.7&9.9&24.44&22.2&11.2&29.75& - & - & - \\
    MPC\textsubscript{Main} + MPC\textsubscript{Synth} &30.4&16.6&47.76&22.2&11.2&29.75&\textbf{95.2} & \textbf{58.7}& \textbf{ 95.66}\\
    GRM &21.8&7.3&20.69&24.6&8.3&23.33& - & -  & - \\
    Seq2Seq LSTM &43.9&14.7&51.43&43.6&12.9&49.16&47.1&29.3&69.18\\
    Seq2Seq Transformer &45.4&16.8&57.50&44.5&14.8&51.79&51.9&32.3&73.62\\
    T5 &48.1&17.4&59.63&46.6&15.2&53.42&59.5&3461&77.18\\
    BART &51.9&18.3&61.89&50.3&16.1&55.64&64.7&35.8&79.55\\
    BART + ITC &50.7&18.3&61.55&49.1&16.0&55.29&63.5&35.9&79.24\\
    BART + MPC\textsubscript{Main}  &53.2&18.6&62.48&51.8&16.4&\textbf{56.58}&64.1&35.6&79.21\\
    \textsc{Trie-NLG} & \textbf{56.5} & \textbf{19.3} & \textbf{66.63} & \textbf{52.0} & \textbf{16.5} & \textbf{56.56} &92.1&41.2&94.62\\
    \hline
    \end{tabular}
    \caption{Results of the models on AOL dataset. Here we report exact evaluation scores, unlike the ones for the Bing dataset. We consider up to 3 completions as additional context from MPC\textsubscript{Main} or MPC\textsubscript{Synth}. GRM is a ranking model based on clicked queries. As the Unseen dataset does not have click information, GRM models cannot be built.
    }
    \label{tab:results_AOL}
\end{table}
\setlength{\tabcolsep}{4pt}

%%%%%
\subsection{Baselines}
Our proposed model is based on both NLG and Trie models. Such joint modeling of NLG systems with popularity signals from trie has not been previously explored. To thoroughly evaluate its performance, we have carefully selected ten diverse baselines, including the basic trie-based model (MPC, MPC+SynthMPC) to ranking (GRM) to Deep learning (LSTM, Transformers) to most recent pre-trained NLG models (T5, BART). In light of the superior performance demonstrated by transformer-based models, we have also included multiple strong transformer-based baselines. As our focus is on generation rather than ranking, we have selected more generative baselines for comparison. However, the outputs of our proposed model can be used as features in learning-to-rank and traditional models. The following baselines have been considered for our experimental evaluation:

%Additionally, As transformer-based models have been shown to outperform traditional approaches, for fairness, we consider multiple strong baselines based on transformers. Additionally, Since we focus on the generation and not ranking, we compare with more generative baselines. However, the proposed model score/outputs can be used with as features in learning-to-rank and traditional models. Considering these observations, we have considered the following baselines: 

%More specifically, there is little existing work with Transformer architectures for QAC. So we also create strong baselines with recent architectures and pre-trained models.

\begin{enumerate}
   \item \textbf{MPC\textsubscript{Train}:} This uses the traditional MPC method~\citep{bar2011context}. The candidate rankings are obtained based on the popularity of each query from the historical query log. Here the historical query log is the training data itself.
   
   \item \textbf{MPC\textsubscript{Main}:} In this baseline the completions are obtained using the main trie created using 1.5 years of historical query logs from Bing. 

   \item \textbf{MPC\textsubscript{Train} + MPC\textsubscript{Synth}}/ \textbf{MPC\textsubscript{Main} + MPC\textsubscript{Synth}}: Completions are obtained using MPC\textsubscript{Train} and MPC\textsubscript{Synth}/MPC\textsubscript{Main} for seen and unseen prefixes resp.
   
%   In this baseline, for seen prefixes the completions are obtained from MPC\textsubscript{Train}. For unseen prefixes  the completions are obtained from MPC\textsubscript{Synth}. Details of MPC\textsubscript{Synth} were presented in Section~\ref{synthmpc}.
   
   \item \textbf{GRM:} We first represent a session, prefixes and complete query as a bag-of-word (BOW) vector and then LambdaMART is trained with these features. %It is also called as the GBDT (Gradient Boosted Decision Trees)-based Ranking model.    
   %\noindent \textbf{DRM:}
   
   \item \textbf{Seq2Seq LSTM:} Standard LSTM based sequence-to-sequence model with attention. Input is a prefix and the target is the complete query.
   
   \item \textbf{Seq2Seq Transformer:} Standard Seq2Seq Transformer model with architecture similar to T5-base. We train the model from scratch. Input is ``session [SEP] prefix'' and the target is the complete query. 
   
   \item \textbf{T5:} Same as Seq2Seq Transformer, except that we \emph{fine-tune} T5-base~\citep{JMLR:v21:20-074} on the QAC dataset.
   
   \item \textbf{BART:} Similar to \cite{mustar2020using}, we fine-tune BART-base \citep{lewis2019bart} with QAC dataset. Input and output are the same as that of T5.
   
   \item \textbf{BART + Implicit Trie Context (ITC):} In this modeling, we try to augment trie's knowledge implicitly. It is a two-step training procedure: (i) BART-base is fine-tuned using a dataset, which consists of session and prefix as input and top $m$ suggestions from MPC\textsubscript{Main}/MPC\textsubscript{Synth} as the target. (ii) This training checkpoint is further trained with the QAC dataset, where the input is the session and the prefix and output are the clicked query. In the second step, we freeze the parameters of the first six decoder layers to retain the trie-based knowledge. During inferencing, the model checkpoint obtained from the second stage of training takes prefix and session as input and outputs the query suggestion.
   
   \item \textbf{BART + MPC\textsubscript{Main}:} This baseline augments the trie knowledge explicitly. With each training example, we add up to top-$m$ trie completions as additional context. There are no completions for unseen prefixes. Input is session, prefix, and additional trie context and output is the clicked query.  
\end{enumerate}

\subsection{Implementation Details}
The proposed model and all the baselines are implemented in Python. GRM is implemented using learning to rank library\footnote{\url{https://github.com/jma127/pyltr}} and seq2seq LSTM is implemented using Texar\footnote{\url{https://github.com/asyml/texar}}. All the transformer-based models are implemented using  HuggingFace Library\footnote{\url{https://huggingface.co/}}. 
All the experiments were conducted on eight A100 Azure cloud GPUs. The batch size is 128; the learning rate is 1e-4, the scheduler is `linear,' the number of epochs is 5, and early stopping was enabled. We used the Adam optimizer with a max source length of 200 and a max target length of 32. BART-base has 6 layers, and 12 heads, layer normalization was enabled and the hidden layer dimension is 768. For the generation, the number of beam size (i.e., $k$) is 8, the maximum sequence length is set to 16, and the repetition penalty\footnote{\url{https://huggingface.co/blog/how-to-generate\#appendix}} is 0.6. We applied grid search for hyper-parameter tuning on the validation dataset and all the scores are reported on the test dataset. We experimented with 1, 3, 5, and 8 as values of $m$, the number of suggestions to extract from the trie. Based on the results of the validation dataset, $m=3$ was selected for running experiments on the test data. We make our code publicly available\footnote{\url{https://github.com/kaushal0494/Trie-NLG}}.

\section{Results and Discussions}
\label{sec:results}
In this section, we present and analyze results of different baselines and the proposed model.

\subsection{Overall Performance Comparison}
\label{sec:resultsOver}
Tables~\ref{tab:results_Bing} and \ref{tab:results_AOL} summarize the experimental results on the Bing and AOL datasets, respectively. Due to the confidential nature of the Bing dataset, we cannot report the exact values of the metrics. This is common practice in many previous studies~\citep{rosset2018optimizing} as well. Hence, in Table~\ref{tab:results_Bing} and the rest of the paper we report percentage improvement scores of the models over reference \textit{MPC\textsubscript{Train} + MPC\textsubscript{Synth}} baseline for the Bing dataset. We cannot use MPC\textsubscript{Train} as a reference for showing percentage improvements as the model does not have any completions for unseen prefixes. For the publicly available AOL dataset, we report exact evaluation scores across all three metrics. We also report percentage improvement scores of the models over reference \textit{MPC\textsubscript{Train} + MPC\textsubscript{Synth}} baseline for the AOL dataset in Table~\ref{tab:results_AOL_relative}. Overall, our proposed \textsc{Trie-NLG} outperforms all the traditional, ranking, and generative models, across both the datasets (including Seen and Unseen) and all three metrics. Paired t-test shows that \textsc{Trie-NLG} outperforms the best baseline statistically significantly across both the datasets for each of the three metrics with a p-value less than 0.05.  

Note that $MPC\textsubscript{Train}+MPC\textsubscript{Synth}$ and $MPC\textsubscript{Main}+MPC\textsubscript{Synth}$ have identical results for ``unseen'' datasets. Similarly, $MPC\textsubscript{Train}$ and $MPC\textsubscript{Train}+MPC\textsubscript{Synth}$ yield same results for ``seen'' dataset. This is expected because $MPC\textsubscript{Train}$ and $MPC\textsubscript{Main}$ provide trie suggestions for seen prefixes; while $MPC\textsubscript{Synth}$ provides trie suggestions for unseen prefixes.

Without MPC\textsubscript{Synth}, the MPC\textsubscript{Train} and MPC\textsubscript{Main} do not have completions for the Unseen dataset. MPC\textsubscript{Synth} provides completions for the unseen prefixes and boosts the overall model performance. As expected, the generative models provide suggestions for unseen prefixes, unlike ranking and database lookup models. Evaluation scores of Seq2Seq Transformer and pre-trained models (i.e., T5 and BART) indicate that the pre-trained models provide better input representation and perform better. BART + ITC fuses the additional context (top-ranked completions obtained from MPC\textsubscript{Main}) implicitly with two-step training. However, the results are not promising, indicating that the model's learning is distracted in the two-stage training. However, adding explicit context leads to better performance, as shown in BART + MPC\textsubscript{Main} model. Eventually, adding context from MPC\textsubscript{Main} and MPC\textsubscript{Synth} helps the proposed \textsc{Trie-NLG} model perform the best.  

The absolute evaluation scores for Unseen AOL data are much higher as compared to Seen AOL data. We observe similar trends for the Bing dataset as well. There could be two possible causes for this: (1) Unseen dataset is $\sim$11\% of the original data, and hence it is much smaller compared to the Seen dataset, and (2) the average prefix lengths for Seen AOL, Unseen AOL, Seen Bing, and Unseen Bing, are  8.1, 20.9, 14.5 and 25.9 respectively. In the Unseen dataset, the lengths of the prefixes are longer compared to Seen, which provides more context and the generative models perform better. Most of the baseline models' performance on the Unseen dataset is very poor,  but the proposed \textsc{Trie-NLG} achieves much better performance which shows the promising prospect of our approach. GRM is a ranking model based on clicked queries. As the Unseen dataset does not have click information, GRM models cannot be built.

The MPCMain + MPCSynth model is the best-performing model for the AOL Unseen dataset, and it surpasses the \textsc{Trie-NLG} by a small margin. However, for the Bing Unseen dataset, the proposed model outperformed all the models. There can be multiple possible reasons behind this observation. For example, (1) \textbf{Dataset Timeline:} The AOL dataset is from 2006 while Bing data was collected in 2020-21. Pre-trained NLG models (like BART and T5) have been trained with recent corpus whose vocabulary is expected to be better aligned with recent Bing data rather than AOL. Final suggestions from the model for Unseen AOL data are hence governed by only the partially-aligned language model and without any context from the trie. (2) \textbf{Query Log Size:} Bing dataset has ~20M queries compared to ~4M in AOL dataset. This leads to better synthetic suggestions for Bing, in turn leading to better context augmentation for the Bing \textsc{Trie-NLG} model. (3) \textbf{Prefix and Session Lengths:} The prefix and session length for Bing (4.434 tokens/prefix and 5.619 queries/session) are longer as compared to AOL (3.061 tokens/prefix and 2.530 queries/session). Longer prefixes and sessions lead to better NLG completions for Bing. Recent search interactions for users do involve longer sessions, and the proposed model is expected to do well in such scenarios.

The overall evaluation results indicate that neither trie nor NLG models are effective individually for such a challenging scenario. The proposed hybrid approach that considers the benefits of both worlds (language semantics from NLG and popularity statistics from trie) through a joint modeling technique is a promising approach and can push the QAC research field forward.

\subsection{Performance Analysis for Short Prefixes}
Table~\ref{tab:short_prefix} shows the performance of our proposed \textsc{Trie-NLG} model, and two baselines BART and BART+MPC\textsubscript{Main} for different prefix lengths on both the datasets. The evaluation scores are reported for three different buckets based on the character length of the prefix: [1-5], [6-10] and 10+. The evaluation scores in 10+ bucket are higher as compared to the other two. It indicates that as the prefix length increases, the performance of all the models increase. It is aligned with the intuition that model generates more accurate predictions as the prefix becomes longer. The model performance improves across both datasets for short prefixes as the additional context is added, i.e., BART+MPC\textsubscript{Main} performs better as compared to BART. Moreover, when synthetic completions are included further, i.e., \textsc{Trie-NLG}, it outperforms both the baselines even for very short prefixes. This provides evidence that adding additional trie knowledge does help to increase relevant context for short prefixes. A few of the $\Delta$ BLEU\textsubscript{RR} scores for Bing are negative for prefix lengths [1-5]. This implies that the performance of the MPC\textsubscript{Main}+MPC\textsubscript{Synth} model is superior to the three models considered, i.e., BART, BART+MPC\textsubscript{Main}, and \textsc{Trie-NLG}. Despite this, the lower negative values for \textsc{Trie-NLG} demonstrate that its performance is better than the other two models. The reason behind this could be attributed to the fact that (1) the MPC\textsubscript{Main}+MPC\textsubscript{Synth} model demonstrates the best performance for the Bing Unseen dataset in terms of unseen prefixes, as discussed in Section \ref{sec:resultsOver}, and (2) the $\Delta$ BLEU\textsubscript{RR} metric takes into account both the BLEU and MRR scores. However, other metrics' results show consistent improvement across all prefix types and both datasets.

\begin{table*}[!htb]
% \captionsetup{font=scriptsize}
  \begin{center}
  \footnotesize
    \begin{tabular}{|l||c|c|c||c|c|c|} 
    \hline
     & \multicolumn{3}{c||}{\textbf{Prefix Length in [1-5]}} & \multicolumn{3}{c|}{\textbf{Prefix Length in [6-10]}}\\ 
    \hline
    \hline
    \textbf{Bing Dataset}&\textbf{$\Delta$MRR} & \textbf{$\Delta$BLEU\textsubscript{RR}} & \textbf{$\Delta$BLEU} & \textbf{$\Delta$MRR} & \textbf{$\Delta$BLEU\textsubscript{RR}} & \textbf{$\Delta$BLEU} \\
       \hline
       BART & 21.6 & -17.6  & 2.9 & 38.5 & 38.4  & 56.8 \\
       BART + MPC\textsubscript{Main} & 52.1 & -16.8  & 10.1 & 55.2 & 49.3 & 72.1  \\
       \textsc{Trie-NLG} & \textbf{53.2} & \textbf{-15.9}  & \textbf{11.4} & \textbf{56.4} & \textbf{50.1} & \textbf{73.6}   \\
       \hline
       \hline
           \textbf{AOL Dataset}&\textbf{MRR} & \textbf{BLEU\textsubscript{RR}} & \textbf{BLEU} & \textbf{MRR} & \textbf{BLEU\textsubscript{RR}} & \textbf{BLEU}  \\
       \hline
       BART &  41.3 & 8.0 & 35.39 & 52.2 & 14.3 & 53.39  \\
       BART + MPC\textsubscript{Main} &  41.5 & 8.0  & 35.25 & 54.1 & 14.7  & 54.08 \\  
       \textsc{Trie-NLG} & \textbf{42.1} & \textbf{8.1}  & \textbf{35.58} & \textbf{55.2} & \textbf{14.9} & \textbf{55.56}  \\
       \hline
    \end{tabular}

        \begin{tabular}{|l||c|c|c|} 
    \hline
     & \multicolumn{3}{c|}{\textbf{Prefix Length 10+}}\\ 
    \hline
    \hline
    \textbf{Bing Dataset}&\textbf{$\Delta$MRR} & \textbf{$\Delta$BLEU\textsubscript{RR}} & \textbf{$\Delta$BLEU}  \\
       \hline
       BART  &49.1 & 190.0  & 120.1 \\
       BART + MPC\textsubscript{Main} & 55.6 & 218.5 &  145.1\\
       \textsc{Trie-NLG} &  \textbf{60.7} & \textbf{221.1 } & \textbf{149.5}   \\
       \hline
       \hline
           \textbf{AOL Dataset}&\textbf{MRR} & \textbf{BLEU\textsubscript{RR}} & \textbf{BLEU}  \\
       \hline
       BART  & 63.2 & 32.8 & 75.40 \\
       BART + MPC\textsubscript{Main} & 65.1 & 33.4  & 76.20\\  
       \textsc{Trie-NLG} &  \textbf{73.4} & \textbf{35.1}  & \textbf{82.79} \\
       \hline
    \end{tabular}
    \caption{Performance analysis for short prefixes. For Bing, \% improvements over MPC\textsubscript{Train}+MPC\textsubscript{Synth} are reported. For AOL, actual scores are reported.}
    \label{tab:short_prefix}
  \end{center}
\end{table*}

%\comr{How to explain/defend the negative $BL_{RR}$ scores for Bing?}

\subsection{Ablation Study}

Table~\ref{tab:ablation_study} shows ablation results. We observe similar trends for both AOL and Bing. In setups 1 to 3, we have removed session and/or external contexts. The model performs worst when both the information are removed (setup 1). Modeling with the only session (setup 2) performs better than a model that uses only trie context (setup 3), which shows the importance of the user's previous search query log. However, setups 4 to 7 that use both information perform even better, indicating the importance of both types of context. Setups 4 to 7 differ from each other in the number of candidate query completions that are used as additional trie context. It is observed that the use of a single top-ranked candidate query results in worse performance, which may be attributed to an inadequate context. Furthermore, incorporating more than three top-ranked queries also results in poor performance. This can be due to two possible factors: (1) the model may become overwhelmed and unable to effectively distinguish relevant information from the trie context in the presence of too many suggestions in the input, or (2) the trie context may become too long, hindering the model's ability to effectively utilize session signals. Overall, \textsc{Trie-NLG} with top-3 trie candidate completions (i.e., $m=3$) in the input performs the best.  

\begin{table*}[!htb]
% \captionsetup{font=scriptsize}
\footnotesize
\centering
    \begin{tabular}{|c|p{1in}||c|c|c||c|c|c|} 
    \hline
    \textbf{\#}& \textbf{Ablation} & \multicolumn{3}{c||}{\textbf{Bing Dataset}} & \multicolumn{3}{c|}{\textbf{AOL Dataset}}\\ 
    \hhline{|~~|------|}
    & \textbf{Criteria} &\textbf{$\Delta$MRR} & \textbf{$\Delta$BLEU\textsubscript{RR}} &  \textbf{$\Delta$BLEU} & \textbf{MRR} & \textbf{BLEU\textsubscript{RR}}  & \textbf{BLEU} \\
      \hline
      \hline
1& No (Trie Context + Session) &-29.5&-1.5&43.1&5.7&5.9&30.9\\
2& No Trie Context &36.7&73.5&91.1&51.9&18.3&61.9\\
3& No Session &29.3&47.3&73.7&15.4&9.5&40.6\\
     4 & \textsc{Trie-NLG}(1) &44.8&81.3&104.8&32.3&15.1&54.4\\
     5 & \textsc{Trie-NLG}(5) &50.9&82.9&110.0&42.6&17.3&59.6\\
     6 & \textsc{Trie-NLG}(8) &52.9&83.8&111.4&28.5&13.8&51.5\\
7& \textsc{Trie-NLG}(3) &\textbf{56.8}&\textbf{88.3}&\textbf{114.5}&\textbf{56.5}&\textbf{19.3}&\textbf{66.6}\\
       \hline
    \end{tabular}
    \caption{Results of ablation study using different experimental setups. \textsc{Trie-NLG}(m) means ``\textsc{Trie-NLG} + Up to Top-m Completions''. For the  Bing dataset, percentage improvements over MPC\textsubscript{Train}+MPC\textsubscript{Synth} baseline are reported. For the AOL dataset, actual evaluation scores are reported.}
    \label{tab:ablation_study}
\end{table*}

\subsection{Trie Completion Retention Analysis}
Further, we  analyze \textit{how many} candidate queries from trie context are generated as completions by the proposed \textsc{Trie-NLG} model, and \textit{what position} they appear in. For simplicity, we only consider up to 3 candidate queries and seen test datasets. In the ideal scenario, the best performing model should retain good candidate queries of MPC\textsubscript{Main} into the recommended completion list as well as generate new completions. Table~\ref{tab:cand_ret1} shows that $\sim$19\% and $\sim$43\% examples do not retain any completions for Bing and AOL datasets, respectively. At the same time, $\sim$23\% of the Bing examples retain all the input candidate queries. For the AOL dataset, only 3\% of examples have all the input candidates; this value is very low because MPC\textsubscript{Main} is created with only Bing historical search log but used for generating completions for AOL prefixes. So the trie-recommended completions may not be very relevant and hence not considered by \textsc{Trie-NLG} for the AOL dataset. 

Tables~\ref{tab:cand_ret2_bing} and~\ref{tab:cand_ret2_aol} provide the position distribution of each trie candidate in the \textsc{Trie-NLG} output for Bing and AOL datasets respectively. In more than 40\% of examples, the top-ranked candidate query from the trie doesn't appear in the final generated output. %In general, the candidate query can appear in any position; sometimes, they dominantly appear in the same position as itself. For instance, 
On the other hand, for 37.6\%  examples, the top trie candidate is also the top suggestion from \textsc{Trie-NLG} for the Bing dataset. This also indicates that the model does not blindly copy the trie candidates as outputs. Instead, it learns to determine the candidate's goodness or fit for the specific input and performs the generation accordingly.

\begin{table}[!htb]
% \captionsetup{font=scriptsize}
  \begin{center}
\footnotesize
    \begin{tabular}{|l|c|c|c|c|} 
    \hline
     $\mathbf{t}$&0&1&2&3\\ 
    \hline
    \hline
\textbf{Bing Seen Test Dataset}&19.0\%&26.4\%&30.5\%&23.6\%\\
\textbf{AOL Seen Test Dataset}&43.1\%&36.1\%&17.2\%&3.2\%\\
       \hline
    \end{tabular}
    \caption{Number of examples where $t$ trie suggestions were retained in the \textsc{Trie-NLG} generated completions for \textit{Seen Test Datasets}.}
    \label{tab:cand_ret1}
  \end{center}
\end{table}

\begin{table*}[!htb]
% \captionsetup{font=scriptsize}
  \begin{center}
\footnotesize
    \begin{tabular}{|l||c|c|c|c|c|c|c|c|c|} 
    \hline
    Rank$\downarrow$/Pos$\rightarrow$ &1 & 2 & 3 & 4 & 5 & 6 & 7 & 8 & None\\
     \hline
     1& 37.60 & 10.17 & 3.92 & 2.10 & 1.43 & 1.15 & 0.94 & 0.97 & 41.69 \\
     2&18.50 & 19.51 & 7.86 & 3.81 & 2.45 & 1.81 & 1.53 & 1.511 & 41.69 \\
     3& 9.98 & 12.50 & 11.82 & 7.04 & 4.49 & 2.74 & 1.88 & 1.67 & 47.83 \\
    \hline
    \end{tabular}
    \caption{Percentage of times the candidate suggestion from trie was copied to [1-8]th positions (`Pos') as output by \textsc{Trie-NLG} for Bing Seen Test Dataset. `None' indicates the candidate suggestion was not a part of \textsc{Trie-NLG} output. Results are shown for Seen Test Data when $m$=3. `Rank' indicates the rank of candidate suggestion from trie.}
    \label{tab:cand_ret2_bing}
  \end{center}
\end{table*}

\begin{table*}[!htb]
% \captionsetup{font=scriptsize}
  \begin{center}
\footnotesize
    \begin{tabular}{|l||c|c|c|c|c|c|c|c|c|} 
    \hline
    Rank$\downarrow$/Pos$\rightarrow$ &1 & 2 & 3 & 4 & 5 & 6 & 7 & 8 & None\\
     \hline
     1&  25.92 & 6.81 & 3.43 & 2.22 & 1.57 & 1.23 & 1.03 & 1.12 & 56.62 \\
     2& 7.41 & 5.47 & 3.26 & 2.23 & 1.65 & 1.29 & 1.11 & 1.11 & 56.62 \\
     3& 3.97 & 3.46 & 2.53 & 1.97 & 1.54 & 1.21 & 1.04 & 1.01 &83.24 \\
    \hline
    \end{tabular}
    \caption{Percentage of times the candidate suggestion from trie was copied to [1-8]th positions (`Pos') as output by \textsc{Trie-NLG} for AOL Seen Test Dataset. `None' indicates the candidate suggestion was not a part of \textsc{Trie-NLG} output. Results are shown for Seen Test Data when $m$=3. `Rank' indicates the rank of candidate suggestion from trie.}
    \label{tab:cand_ret2_aol}
  \end{center}
\end{table*}

% In fact, in quite a few instances %MPC\textsubscript{Main} generates correct completions at the first position, and 
%  \textsc{Trie-NLG} retains the topmost suggestion from the trie at the first position. This provides evidence that \textsc{Trie-NLG} model learns to understand good and bad candidate query completions and makes the decision accordingly to provide the best recommendations.   

\subsection{Runtime Analysis}
Table~\ref{tab:runtime} shows execution runtimes for the three models on an A100 Nvidia GPUs. Trie lookups are very cheap compared to BART-based suggestion generation. Hence, our method \textsc{Trie-NLG} has almost similar runtimes compared to a standard BART model.

 \setlength{\tabcolsep}{2pt}
\begin{table}[!htb]
% \captionsetup{font=scriptsize}
\footnotesize
\centering
    \begin{tabular}{|p{0.90in}||c|c|c||c|c|c|} 
    \hline
    \textbf{Models} & \multicolumn{3}{c||}{\textbf{Bing Test Dataset}} & \multicolumn{3}{c|}{\textbf{AOL Test Dataset}}\\ 
    \hhline{|~|------|}
     \textbf{} &\textbf{Seen} & \textbf{Unseen} &  \textbf{Total} & \textbf{Seen} & \textbf{Unseen}  & \textbf{Total} \\
      \hline
      \hline
 BART &11.25&11.75&11.29&11.69&12.83&11.82\\
 BART+ MPC\textsubscript{Main} &12.28&12.16&12.27&12.17&13.05&12.27\\
 TRIE-NLG &12.34&12.25&12.34&12.29&13.20&12.40\\
       \hline
    \end{tabular}
    \caption{Runtime of different test dataset splits. Values are in millisecond(ms)/record for 8 auto-complete generations.}
    \label{tab:runtime}
\end{table}
\setlength{\tabcolsep}{4pt}

\subsection{Case Studies}
\begin{figure*}[!htb]
% \captionsetup{font=scriptsize}
\centering
\includegraphics[width=0.9\textwidth]{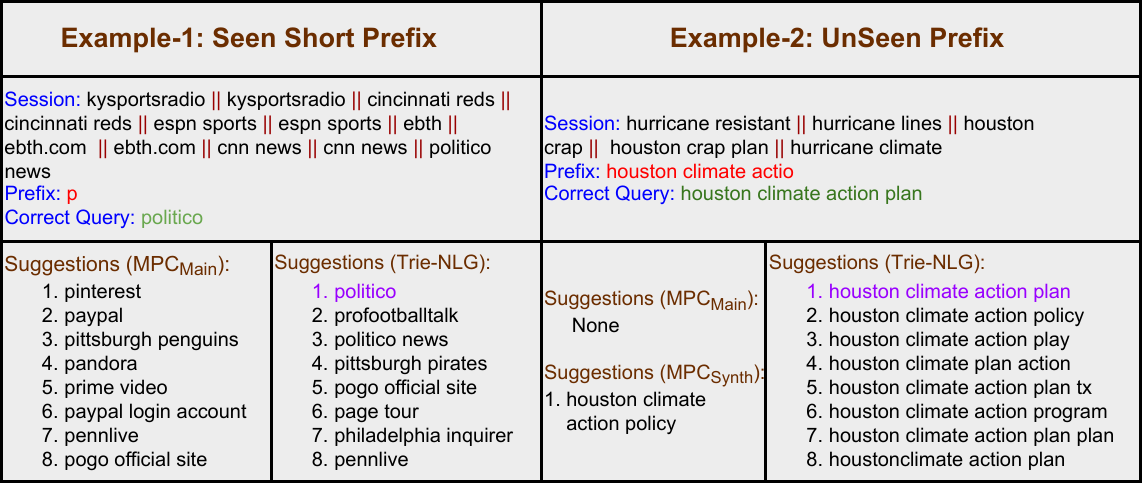}
\caption{Generated query completions for seen (short) and unseen prefixes.} 
\label{fig:case_study}
\end{figure*}

Fig.~\ref{fig:case_study} shows two examples of suggestions for a short seen prefix and an unseen prefix respectively. In the first example, the prefix `p' is very short and MPC\textsubscript{Main} is unable to understand the context and recommends more general/popular completions. Whereas the proposed \textsc{Trie-NLG} model learned the personalized context and recommended correct and more relevant completions. The model also considers recommendations from MPC\textsubscript{Main} as additional context. For instance, `pogo official site' is present in MPC\textsubscript{Main} and recommended by \textsc{Trie-NLG}, although there is no relevant context in the session. In Example-2, there is no query recommendation from MPC\textsubscript{Main} for the unseen prefix, but MPC\textsubscript{Synth} has one recommendation and that acts as additional context for \textsc{Trie-NLG}. \textsc{Trie-NLG} generates more relevant completions and the top-ranked completion is correct. In summary, we can conclude that the additional trie context is useful for the generative model and helps \textsc{Trie-NLG} to generate more accurate and relevant query completions.

\section{Conclusion}
\label{sec:conclusions}

We proposed \textsc{Trie-NLG} model for personalized QAC. It is based on context augmentation in the NLG model where the additional context is obtained from the main trie or the synthetic trie. To the best of our knowledge, this is the first study to use the trie context in NLG models for QAC. We primarily focused on solving the problem of short and unseen prefixes. The model was evaluated on a prepared AOL QAC dataset and a real prefix-to-click QAC dataset from Bing. The proposed model outperformed all the baselines while specifically improving the performance for short and unseen prefixes.

\section{Future Work and Limitations}
\label{sec:future_works}
\textbf{Future Work:} Not all session queries are relevant to the current user prefix. Irrelevant session queries leads to noisy training data. In the future, we plan to explore modeling techniques that can select and encode only the relevant queries as personalized contexts for \textsc{Trie-NLG}. Additionally, we will explore \textit{on-the-fly} models like RAG \citep{lewis2020retrieval}, which can provide additional (better) content or completions instead of trie completions to boost the performance. Finally, we will explore a transfer learning approach to extend this to multilingual QAC system.

\noindent\textbf{Limitations:} The proposed \textsc{Trie-NLG} model operates in a two-step process, comprising the extraction of auto-completions from a trie and augmentation in the NLG model. As a consequence of this approach, the model exhibits slightly higher latency compared to the standard NLG model due to trie lookup. However, the trie lookup time is notably low. It is crucial to highlight that the proposed model is built upon a pre-trained NLG model (BART), which renders it susceptible to displaying unexpected outcomes inherited from the pre-training phase \citep{gehman-etal-2020-realtoxicityprompts}. Such outcomes may include toxicity, bias, hallucination, misinformation, and other similar issues. %We will explore the safety concerns in the future.}

\backmatter
\section*{Declarations}
\begin{itemize}
\item Funding

This research was partially funded by Microsoft Academic Partnership Grant 2022.

\item Competing interests 

The authors have no competing interests to declare relevant to this article's content. 

% \item Ethics approval

\item Consent for publication

All the authors of this work have been informed of the submission, and everyone supports it. 

\item Availability of data and materials

We make our code available as supplementary information.  It is in line with our claims in the paper and can be trusted to reproduce the results. We use one public dataset (AOL) and a Bing dataset. Since Bing data is proprietary, it cannot be made publicly available.

\item Code availability 

We make our code publicly available and it is attached as supplementary data. 

\item Authors' contributions

All authors contributed to the design of the project. Specifically,  Conceptualization, Methodology, Data preparation: [Kaushal Kumar Maurya]; Writing - original draft preparation: [Kaushal Kumar Maurya, Manish Gupta, Maunendra Sankar Desarkar]; Writing - review and editing: [everyone]; Funding acquisition, Resources, Supervision: [Manish Gupta, Maunendra Sankar Desarkar, Puneet Agrawal].

\end{itemize}

\bibliography{references}

\end{document}